\documentclass[table]{bmcart}

\usepackage[utf8]{inputenc} 
\usepackage[T1]{fontenc}
\usepackage{arydshln}
\usepackage{caption}
\usepackage{amsmath} 
\usepackage{verbatim}
\usepackage{setspace}
\usepackage{amsfonts}
\usepackage{lineno}
\usepackage{mathtools}
\usepackage{xcolor}
\usepackage{soul}
\usepackage{hyperref}
\usepackage{pdfpages}
\usepackage{gensymb}
\usepackage[utf8]{inputenc}
\usepackage{adjustbox}
\usepackage{chngpage}
\usepackage{comment}
\usepackage{float}
\usepackage{comment}
\usepackage[bordercolor=white,backgroundcolor=gray!30,linecolor=black,colorinlistoftodos]{todonotes}

\startlocaldefs
\endlocaldefs

\begin{document}

\begin{frontmatter}

\begin{fmbox}
\dochead{Research}


\title{Automated freezing of gait assessment with marker-based motion capture and multi-stage spatial-temporal graph convolutional neural networks}


\author[
   addressref={aff1,aff2},                   
   corref={aff1},                       
   email={benjamin.filtjens@kuleuven.be}   
]{\fnm{Benjamin} \snm{Filtjens}}
\author[
   addressref={aff3},
]{\fnm{Pieter} \snm{Ginis}}
\author[
   addressref={aff3},
]{\fnm{Alice} \snm{Nieuwboer}}
\author[
   addressref={aff2},
]{\fnm{Peter} \snm{Slaets}}
\author[
   addressref={aff1},
]{\fnm{Bart} \snm{Vanrumste}}


\address[id=aff1]{%
  \orgname{eMedia Research Lab/STADIUS, Department of Electrical Engineering (ESAT), KU Leuven}, 
  \street{Andreas Vesaliusstraat 13},                     %
  \postcode{3000}                                
  \city{Leuven},                              
  \cny{Belgium} 
}
\address[id=aff2]{
  \orgname{Intelligent Mobile Platforms Research Group, Department of Mechanical Engineering, KU Leuven},
  \street{Andreas Vesaliusstraat 13},
  \postcode{3000}
  \city{Leuven},
  \cny{Belgium}                                 
}
\address[id=aff3]{%
  \orgname{Research Group for Neurorehabilitation (eNRGy), Department of Rehabilitation Sciences, KU Leuven},
  \street{Tervuursevest 101},
  \postcode{3001}
  \city{Heverlee},
  \cny{Belgium}
}


\begin{artnotes}
\end{artnotes}

\end{fmbox}


\begin{abstractbox}

\begin{abstract} 
\parttitle{Background} 
Freezing of gait (FOG) is a common and debilitating gait impairment in Parkinson's disease. Further insight into this phenomenon is hampered by the difficulty to objectively assess FOG. To  meet  this  clinical  need,  this  paper  proposes  an automated motion-capture-based  FOG  assessment  method  driven  by  a  novel  deep neural network. 
\parttitle{Methods} 
Automated FOG assessment can be formulated as an action segmentation problem, where temporal models are tasked to recognize and temporally localize the FOG segments in untrimmed motion capture trials. This paper takes a closer look at the performance of state-of-the-art action segmentation models when tasked to automatically assess FOG. Furthermore, a novel deep neural network architecture is proposed that aims to better capture the spatial and temporal dependencies than the state-of-the-art baselines. The proposed network, termed multi-stage spatial-temporal graph convolutional network (MS-GCN), combines the spatial-temporal graph convolutional network (ST-GCN) and the multi-stage temporal convolutional network (MS-TCN). The ST-GCN captures the hierarchical spatial-temporal motion among the joints inherent to motion capture, while the multi-stage component reduces over-segmentation errors by refining the predictions over multiple stages. The proposed model was validated on a dataset of fourteen freezers, fourteen non-freezers, and fourteen healthy control subjects. 
\parttitle{Results} 
The experiments indicate that the proposed model outperforms four state-of-the-art baselines. Moreover, FOG outcomes derived from MS-GCN predictions had an excellent (r=0.93 [0.87, 0.97]) and moderately strong (r=0.75 [0.55, 0.87]) linear relationship with FOG outcomes derived from manual annotations.
\parttitle{Conclusions} 
The proposed MS-GCN may provide an automated and objective alternative to labor-intensive clinician-based FOG assessment. Future work is now possible that aims to assess the generalization of MS-GCN to a larger and more varied verification cohort.
\end{abstract}




\end{abstractbox}
%

\end{frontmatter}



\section*{Background}
Freezing of gait (FOG) is a common and debilitating gait impairment of PD. Up to 80\% of people with Parkinson's disease (PwPD) may develop FOG during the course of the disease \cite{Perez-Lloret2014-xr, Hely2008-rw}. FOG leads to sudden blocks in walking and is clinically defined as a "brief,  episodic  absence  or  marked  reduction  of  forward  progression  of  the  feet  despite  the  intention  to  walk and reach a destination” \cite{Nutt2011-we}. The PwPD themselves describe freezing of gait as “the feeling that their feet are glued to the ground” \cite{Snijders2008-vt}. Freezing episodes most frequently occur while traversing under environmental constraints, during emotional stress, during cognitive overload by means of dual-tasking, and when initiating gait \cite{Nonnekes2015-ec, Okuma2014-tl}. Though, turning hesitation was found to be the most frequent trigger of FOG \cite{Schaafsma2003-pz, Giladi2002-bf}. Subjects with FOG experience more anxiety \cite{Giladi2006-gu}, have a lower quality of life \cite{Moore2005-ph}, and are at a much higher risk of falls \cite{Bloem2004-na, Grimbergen2004-fk, Gray2000-fh, Rudzinska2013-mo, Pelicioni2019-tt}. \\ 
Given the severe adverse effects associated with FOG, there is a large incentive to advance novel interventions for FOG \cite{Gilat2018-di}. Unfortunately, the pathophysiology of FOG is complex and the development of novel treatments is severely limited by the difficulty to objectively assess FOG \cite{Mancini2019-sn}. Due to heightened levels of attention, it is difficult to elicit FOG in the gait laboratory or clinical setting \cite{Okuma2014-tl, Snijders2008-vt}. Therefore, health professionals relied on subjects' answers to subjective self-assessment questionnaires \cite{Giladi2000-hj, Nieuwboer2009-ca}, which may be insufficiently reliable to detect FOG severity \cite{Shine2012-mw}. Visual analysis of regular RGB videos has been put forward as the gold standard for rating FOG severity \cite{Shine2012-mw, Gilat2019-xm}. However, the visual analysis relies on labor-intensive manual annotation by a trained clinical expert. As a result, there is a clear need for an automated and objective approach to assess FOG.\\
The percentage time spent frozen (\%TF), defined as the cumulative duration of all FOG episodes divided by the total duration of the walking task, and the number of FOG episodes (\#FOG) have been put forward as reliable outcome measures to objectively assess FOG \cite{Morris2012-hl}. An accurate segmentation in-time of the FOG episodes, with minimal over-segmentation errors, is required to robustly determine both outcome measures. \\ 
Several methods have been proposed for automated FOG assessment based on motion capture (MoCap) data. MoCap encodes human movement as a time series of human joint locations and orientations or their higher-order representations and is typically performed with optical or inertial measurement systems. Prior work has tackled automated FOG assessment as an action recognition problem and used a sliding-window scheme to segment a MoCap sequence into fixed partitions \cite{Moore2008-hz,Moore2013-ns, Popovic2010-jq,Delval2010-ck,Hu2020-kq,Ahlrichs2016-ay,Rodriguez-Martin2017-uf,Masiala2019-kz,Tahafchi2017-pd,Camps2017-ji,Sigcha2020-ft,Mancini2012-et, Mancini2021-wn, ODay2021-ls}. For all the samples within a partition, a single label is then predicted with methods ranging from simple thresholding methods \cite{Moore2008-hz,Delval2010-ck} to high-level temporal models driven by deep learning \cite{Hu2020-kq,Masiala2019-kz,Camps2017-ji,Sigcha2020-ft, ODay2021-ls}. However, the samples within a pre-defined partition may not always share the same label. Therefore, a data-dependent heuristic is imposed to force all samples to take a single label, most commonly by majority voting \cite{Sigcha2020-ft, ODay2021-ls}. Moreover, a second data-dependent heuristic is needed to define the duration of the sliding-window, which is a trade-off between expressivity, i.e. the ability to capture long-term temporal patterns, and sensitivity, i.e. the ability to identify short-duration FOG episodes. Such manually defined heuristics are unlikely to generalize across study protocols. \\
This study proposes to reformulate the problem of FOG annotation as an action segmentation problem. Action segmentation approaches overcome the need for manually defined heuristics by generating a prediction for each sample within a long untrimmed MoCap sequence. Several methods have been proposed to tackle action segmentation. Similar to FOG assessment, earlier studies made use of sliding-window classifiers \cite{Rohrbach2012-av, Ni2016-yx}, which do not capture long-term temporal patterns \cite{Lea2016-yw}. Other approaches use temporal models such as hidden Markov models \cite{Kuehne2015-jp, Tang2012-vf} and recurrent neural networks \cite{Singh2016-wa, Huang2016-gf}. The state-of-the-art methods tend to use temporal convolutional neural networks (TCN), which have been shown to outperform recurrent methods \cite{Lea2016-yw, Bai2018-jk}. Dilation is frequently added to capture long-term temporal patterns by expanding the temporal receptive field of the TCN models \cite{Yu2015-qu}. In multi-stage temporal convolutional network (MS-TCN), the authors show that multiple stages of temporal dilated convolutions significantly reduce over-segmentation errors \cite{Farha2019-yw}. These action segmentation methods have historically been validated on video-based datasets \cite{Fathi2011-kf, Stein2013-od} and thus employ video-based features \cite{Carreira2017-bb}. The human skeleton structure that is inherent to MoCap has thus not been exploited by prior work in action segmentation. \\ 
To model the structured information among the markers, this paper uses the spatial-temporal graph convolutional neural network (ST-GCN) \cite{Yan2018-jp} as the first stage of an MS-TCN network. ST-GCN applies spatial graph convolutions on the human skeleton graph at each time step and applies dilated temporal convolutions on the temporal edges that connect the same markers across consecutive time steps. The proposed model, termed multi-stage spatial-temporal graph convolutional neural network (MS-GCN), thus extends MS-TCN to skeleton-based data for enhanced action segmentation within MoCap sequences. \\
The MS-GCN was tasked to recognize and localize FOG segments in a MoCap sequence. The predicted segments were quantitatively and qualitatively assessed versus the agreed-upon annotations by two clinical-expert raters. From the predicted segments, two clinically relevant FOG outcomes, the \%TF and \#FOG, were computed and statistically validated. To the best of our knowledge, the proposed MS-GCN is a novel neural network architecture for skeleton-based action segmentation in general and FOG segmentation in particular. The benefit of MS-GCN for FOG assessment is four-fold: (1) It exploits ST-GCN to model the structured information inherent to MoCap. (2) It allows modeling of long-term temporal context to capture the complex dynamics that precede and succeed FOG. (3) It can operate on high temporal resolutions for fine-grained FOG segmentation with precise temporal boundaries. (4) To accomplish (2) and (3) with minimal over-segmentation errors, MS-GCN utilizes multiple stages of refinements.

\section*{Methods}
\subsection*{Dataset}
Two existing MoCap datasets \cite{Spildooren2010-pj, Vervoort2016-zm} were included for analysis. The first dataset \cite{Spildooren2010-pj}, includes forty-two subjects. Twenty-eight of the subjects were diagnosed with PD by a movement disorders neurologist. Fourteen of the PwPD were classified as freezers based on the first question of the New Freezing of Gait Questionnaire (NFOG-Q): ``Did you experience ``freezing episodes'' over the past month?'' \cite{Nieuwboer2009-ca}. The remaining fourteen subjects were age-matched healthy controls. The second dataset \cite{Vervoort2016-zm}, includes seventeen PwPD and FOG, as classified by the NFOG-Q. The subjects underwent a gait assessment at baseline and after twelve months follow-up. Five subjects only underwent baseline assessment and four subjects dropped out during the follow-up. The clinical characteristics are presented in Table \ref{tab:subjner}.

\begin{table}[H]
\renewcommand{\arraystretch}{1.2}
\caption[Subject characteristics.]{Left of the vertical line denotes the subject characteristics of the fourteen healthy control subjects (controls), fourteen PwPD and without FOG (non-freezers), and fourteen PwPD and FOG (freezers) of dataset 1. The right of the vertical line denotes the subject characteristics of the seventeen PwPD and FOG (freezers) of dataset 2 at the baseline assessment. All characteristics are given in terms of mean $\pm$ standard deviation. For dataset 1, the characteristics were measured during the ON-phase of the medication cycle, while for dataset 2 the characteristics were measured while OFF medication.}
\label{tab:subjner}
\centering
\begin{tabular}{llll|l}
\\
\hline
                         & Controls   & Non-freezers & Freezers & Freezers  \\ \hline
Age & 65 $\pm$ 6.8 & 67 $\pm$ 7.4   & 69 $\pm$ 7.4 & 67 $\pm$ 9.3 \\
PD duration &  & 7.8 $\pm$ 4.8    & 9.0 $\pm$ 4.8 & 10 $\pm$ 6.3 \\
MMSE \cite{Folstein1975-xf}      &  29 $\pm$ 1.3    & 29 $\pm$ 1.2   & 28 $\pm$ 1.1 & 28 $\pm$ 1.3 \\
UPDRS III \cite{Goetz2008-xh}  &    & 34 $\pm$ 9.9   & 38 $\pm$ 14 & 39 $\pm$ 12 \\
H\&Y \cite{Hoehn1967-ui}  &     & 2.4 $\pm$ 0.3    & 2.5 $\pm$ 0.5 & 2.4 $\pm$ 0.5  \\ \hline
\end{tabular}
\end{table}

\subsection*{Protocol}
\begin{figure*}[!t]
\centering
\includegraphics[width=3.5in]{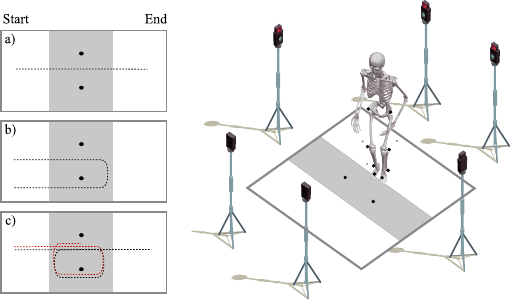}
\caption[Overview of the acquisition protocol.]{Overview of the acquisition protocol. Two reflective markers were placed in the middle of the walkway at a .5m distance from each other to demarcate the turning radius. The data collection included straight-line walking (a), 180 degrees turning (b), and 360 degrees turning (c). The protocol was standardized by demarcating a zone of 1m before and 1m after the turn in which data was collected. The grey shaded area visualizes the data collection zone, while the dashed lines indicate the trajectory walked by the subjects. For dataset 2, the data collection only included straight-line walking and 360-degree turning. Furthermore, the data collection ended as soon as the subject completed the turn, as visualized by the red dashed line.}
  \label{fig_protocol}
\end{figure*}

Both datasets were recorded with a Vicon 3D motion analysis system recording at a sampling frequency of 100 Hz. Retro-reflective markers were placed on anatomical landmarks according to the full-body or lower-limb plug-in-gait model \cite{Kadaba1990-cw, Davis1991-tr}. Both datasets featured a nearly identical standardized gait assessment protocol, where two retro-reflective markers placed .5 m from each other indicated where subjects either had to walk straight ahead, turn 360\degree{}left, or turn 360\degree{}right. For dataset 1, the subjects were additionally instructed to turn 180\degree{}left and turn 180\degree{}right. The experimental conditions were offered randomly and performed with or without a verbal cognitive dual-task \cite{Canning2006-dj,Bowen2001-uc}. All gait assessments were conducted during the off-state of the subjects' medication cycle, i.e. after an overnight withdrawal of their normal medication intake. The experimental conditions are visualized in Figure \ref{fig_protocol}. \\
For dataset 1, two clinical experts, blinded for NFOG-Q score, annotated all FOG episodes by visual inspection of the knee-angle data (flexion-extension) in combination with the MoCap 3D images. For dataset 2, the FOG episodes were annotated by one of the authors (BF) based on visual inspection of the MoCap 3D images. To ensure that the results were unbiased, the FOG trials of dataset 2 were used to enrich the training dataset and not for the evaluation of the model. For both datasets, the onset of FOG was determined at the heel strike event prior to delayed knee flexion. The termination of FOG was determined at the foot-off event that is succeeded by at least two consecutive movement cycles \cite{Spildooren2010-pj}.

\subsection*{FOG segmentation}
Marker-based optical MoCap describes the 3D movement of optical markers in time, where each marker represents the 3D coordinates of the corresponding anatomical landmark. The duration of a MoCap trial can vary substantially due to high inter-and intra-subject variability. The goal is to segment a FOG episode in time, given a variable-length MoCap trial. The MoCap trial can be represented as $X \in \mathbb{R} ^ {N \times T \times C_{in}}$, where $N$ specifies the number of optical markers, $T$ the number of samples, and $C_{in}$ the feature dimension. Each MoCap trial $X$ is associated with a ground truth label vector $Y_{exp}^{T \times l}$, where the label $l$ represents the manual annotation of FOG and functional gait (FG) by the clinical experts. A deep neural network segments a FOG episode in time by learning a function $f: X \rightarrow Y$ that transforms a given input sequence $X = x_{0}, \dots, x_{T}$ into an output sequence $\hat{Y} = \hat{y}_{0}, \dots, \hat{y}_{T}$ that closely resembles the manual annotations $Y_{exp}$. \\
From the 3D marker coordinates, the marker displacement between two consecutive samples was computed as $X(n, t+1, :) - X(n, t, :)$. The two markers on the femur and tibia, which were wand markers in dataset 1 and thus placed away from the primary axis, were excluded. The heel marker was excluded due to close proximity with the ankle marker. The reduced marker configuration consists of nine optical markers. As a result, an input sequence $X \in \mathbb{R} ^ {N \times T \times C_{in}}$ is composed of nine optical markers ($N$), variable duration ($T$), and with the feature dimension ($C_{in}$) composed of the 3D displacement of each marker.

\subsubsection*{MS-GCN}
\begin{figure*}[!t]
\centering
\includegraphics[width=2.5in]{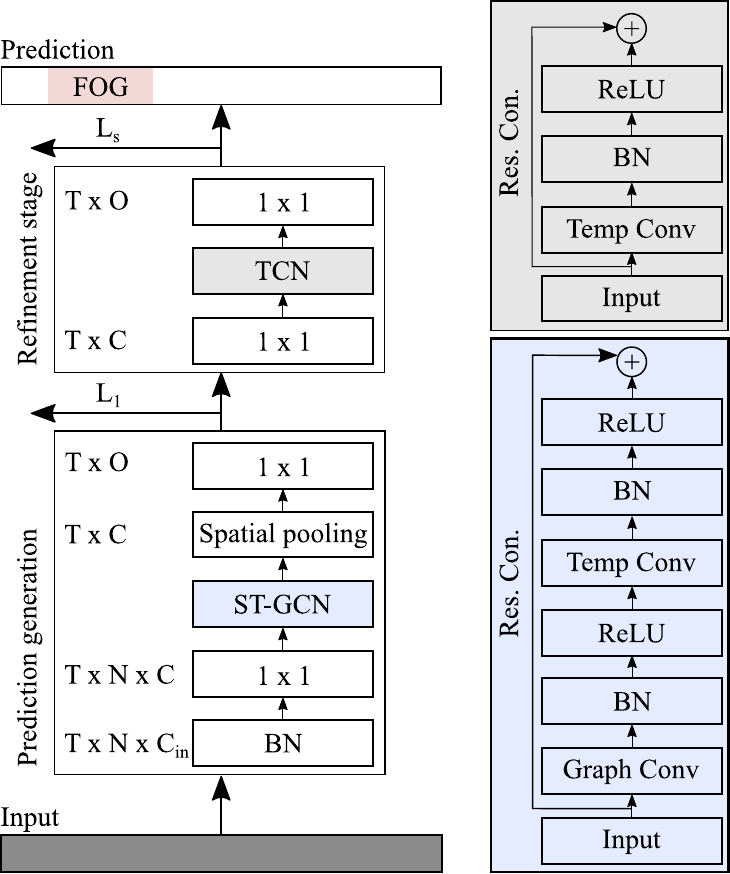}
\caption[Overview of the multi-stage graph convolutional neural network architecture (MS-GCN).]{Overview of the multi-stage graph convolutional neural network architecture (MS-GCN). MS-GCN generates an initial prediction with multiple blocks of spatial-temporal graph convolutional neural network (ST-GCN) layers and refines the predictions over several stages with multiple blocks of temporal convolutional (TCN) layers. An ST-GCN block is visualized in blue and a TCN block in grey.}
\label{fig_model}
\end{figure*}

The proposed multi-stage graph convolutional neural network (MS-GCN), generalizes the multi-stage temporal convolutional neural network (MS-TCN) \cite{Farha2019-yw} to graph-based data. A visual overview of the model architecture is provided in Figure \ref{fig_model}.\\
Formally, MS-GCN features a prediction generation stage of several ST-GCN blocks, which generates an initial prediction $Y \in \mathbb{R}^{T\times l}$. The first layer of the prediction generation stage is a batch normalization (BN) layer that normalizes the inputs and accelerates training \cite{Ioffe2015-ta}. The normalized input is passed through a $1 \times 1$ convolutional layer that adjusts the input dimension $C_{in}$ to the number of filters $C$ in the network, formalized as:
\begin{equation}
f_{adj} = W_1*f_{in}+b,
\end{equation}
where $f_{adj} \in \mathbb{R}^{T\times N\times C}$ is the adjusted feature map, $f_{in} \in \mathbb{R}^{T\times N\times C_{in}}$ the input MoCap sequence, $b \in \mathbb{R}^{C}$ the bias term, $*$ the convolution operator, $W_1 \in \mathbb{R}^{1\times 1\times C_{in}\times C}$ the weights of the $1\times1$ convolution filter with $C_{in}$ input feature channels and $C$ equal to the number of feature channels in the network. \\
The adjusted input is passed through several blocks of ST-GCN \cite{Yan2018-jp}. Each ST-GCN first applies a graph convolution, formalized as:
\begin{equation}
f_{gcn} = \sum_{p} A_p f_{adj}W_p M_p,
\end{equation}
where $f_{adj} \in \mathbb{R}^{T \times N \times C}$ is the adjusted input feature map, $f_{gcn} \in \mathbb{R}^{T \times N \times C}$ the output feature map of the spatial graph convolution, and $W_p$ the $1 \times 1 \times C \times C$ weight matrix. The matrix ${A_p} \in \{0,1\}^{N\times N}$ is the adjacency matrix, which represents the spatial connection between the joints. The graph is partitioned into three subsets based on the spatial partitioning strategy \cite{Yan2018-jp}. The matrix $M_p$ is a learnable ${N\times N}$ attention mask that indicates the importance of each node and its spatial partitions. \\
Next, after passing through a BN layer and ReLu non-linearity, the ST-GCN block performs a dilated temporal convolution \cite{Yu2015-qu}. The dilated temporal convolution is, in turn, passed through a BN layer and ReLU non-linearity, and lastly, a residual connection is added between the activation map and the input. This process is formalized as:
\begin{equation}
f_{out} = \delta(BN(W*_d f_{gcn}+b)) + f_{adj},
\end{equation}
where $f_{out} \in \mathbb{R}^{T\times N\times C}$ is the output feature map, $b \in \mathbb{R}^{C}$ the bias term, $*_d$ the dilated convolution operator, $W \in \mathbb{R}^{k \times 1\times C\times C}$ the weights of the dilated convolution filter with kernel size $k$. The output feature map is passed through a spatial pooling layer that aggregates the spatial features among the $N$ joints. \\
Lastly, the aggregated feature map is passed through a $1 \times 1$ convolution and a softmax activation function to get the probabilities for the $l$ output classes for each sample in-time, formalized as:
\begin{equation}
\hat{y}_{t} = \zeta(W_1 * f_{out} + b),
\end{equation}
where $\hat{y}_{t}$ are the class probabilities at time $t$, $f_{out}$ the output of the pooled ST-GCN block at time $t$, $b \in \mathbb{R}^{l}$ the bias term, $*$ the convolution operator, $\zeta$ the softmax function, $W_1 \in \mathbb{R}^{1\times C \times l}$ the weights of the $1\times 1$ convolution filter with $C$ input channels and $l$ output classes. \\
Next, the initial prediction is passed through one or more refinement stages. The first layer of the refinement stage is a $1 \times 1$ convolutional layer that adjusts the input dimension $l$ to the number of filters $C$ in the network, formalized as:
\begin{equation}
f_{adj} = W_1*f_{in}+b,
\end{equation}
where $f_{adj} \in \mathbb{R}^{T\times C}$ is the adjusted feature map, $f_{in} \in \mathbb{R}^{T\times l}$ the softmax probabilities of the previous stage, $b \in \mathbb{R}^{C}$ the bias term, $*$ the convolution operator, $W_1 \in \mathbb{R}^{1\times l \times C}$ the weights of the $1\times 1$ convolution filter with $l$ input feature channels and $C$ equal to the number of feature channels in the network. \\
The adjusted input is passed through ten blocks of TCN. Each TCN block applies a dilated temporal convolution \cite{Yu2015-qu}, BN, ReLU non-linear activation, and a residual connection between the activation map and the input. Formally, this process is defined as:
\begin{equation}
f_{out} = \delta(BN(W*_df_{adj}+b)) + f_{adj},
\end{equation}
where $f_{out} \in \mathbb{R}^{T\times C}$ is the output feature map, $b \in \mathbb{R}^{C}$ the bias term, $*_d$ the dilated convolution operator, $W \in \mathbb{R}^{k\times C\times C}$ the weights of the dilated convolution filter with kernel size $k$, and $\delta$ the ReLU function. \\
Lastly, the feature map is passed through a $1 \times 1$ convolution and a softmax activation function to get the probabilities for the $l$ output classes for each sample in-time, formalized as:
\begin{equation}
\hat{y}_{t} = \zeta(W_1 * f_{out} + b),
\end{equation}
where $\hat{y}_{t}$ are the class probabilities at time $t$, $f_{out}$ the output of the last TCN block at time $t$, $b \in \mathbb{R}^{l}$ the bias term, $*$ the convolution operator, $\zeta$ the softmax function, $W_1 \in \mathbb{R}^{1\times C \times l}$ the weights of the $1\times 1$ convolution filter with $C$ input channels and $l$ output classes.

\subsubsection*{Model comparison}
To put the MS-GCN results into context, four strong DL baselines were included. Specifically, the state-of-the-art in skeleton-based action recognition, spatial-temporal graph convolutional network (ST-GCN) \cite{Yan2018-jp}. The state-of-the-art in action segmentation, multi-stage temporal convolutional neural network (MS-TCN) \cite{Farha2019-yw}. Two commonly used sequence to sequence models in human movement analysis \cite{Filtjens2020-hl, Matsushita2021-kx}, a bidirectional long short term memory-based network (LSTM) \cite{Graves2005-gv}, and a temporal convolutional neural network-based network (TCN) \cite{Lea2016-yw}. 

\subsubsection*{Implementation details}
To train the models, this paper used the same loss as MS-TCN which utilized a combination of a classification loss (cross-entropy) and smoothing loss (mean squared error) for each stage. The combined loss is defined as:
\begin{equation}
L = L_{cls} + \lambda L_{T-MSE},
\end{equation}
where the hyperparameter $\lambda$ controls the contribution of each loss function. The classification loss $L_{cls}$ is the cross entropy loss:
\begin{equation}
L_{cls} = \frac{1}{T} \sum_t -y_{t,l} log(\hat{y}_{t,l}).
\end{equation}
The smoothing loss $L_{T-MSE}$ is a truncated mean squared error of the sample-wise log-probabilities:
\begin{equation}
L_{T-MSE} = \frac{1}{TC} \sum_{t,c} \widetilde{\Delta}_{t,c}^2
\end{equation}
\[
\widetilde{\Delta}_t =
\begin{cases}
\Delta_{t,c} & \text{: } \Delta_{t,c} \leq \tau,\\
\tau  & \text{: } \hfill\text{otherwise},
\end{cases}
\]
\[
\Delta_{t,l}=|log(\hat{y}_{t,l})-log(\hat{y}_{t-1,l})|,
\]
In each loss function, $T$ are the number of samples and $\hat{y}_{t,l}$ is the probability of FOG or FG at sample $t$. To train the entire network, the sum of the losses over all stages is minimized:
\begin{equation}
L = \sum_{s} L_s
\end{equation}
\\
To allow an unbiased comparison, the model and optimizer hyperparameters were selected according to MS-TCN \cite{Farha2019-yw}. Specifically, the multi-stage models had 1 prediction generation stage and 4 refinement stages. Each stage had 10 layers of 64 filters that applied graph and/or dilated temporal convolutions with kernel size 3 and ReLU activations. The temporal convolutions were acausal, i.e. they could take into account both past and future input features, with a dilation factor that doubled at each layer, i.e. 1, 2, 4, ..., 512. The single-stage models, i.e. ST-GCN and TCN, used the same configuration but without refinement stages. The Bi-LSTM used a configuration that is conventional in human movement analysis, with two forward LSTM layers and two backward LSTM layers, each with 64 cells \cite{Kidzinski2019-ou, Matsushita2021-kx}. For the loss function, $\tau$ was set to 4 and $\lambda$ was set to 0.15. All experiments used the Adam optimizer \cite{Kingma2014-va} with a learning rate of 0.0005. All models were trained for 100 epochs with a batch size of 16. \\
For the temporal models, i.e. LSTM, TCN, and MS-TCN, the input is reshaped into their accepted formats. Specifically, the data is shaped into $T \times C_{in}*N$, i.e. the spatial feature dimension $N$ is thus collapsed. \\
The LSTM was additionally evaluated as an action recognition model. For this evaluation, the MoCap sequences were partitioned into two-second windows and majority voting was used to force all samples to take a single label. These settings are commonly used in FOG recognition \cite{Sigcha2020-ft, ODay2021-ls}. The last hidden LSTM state, which constitutes a compressed representation of the entire sequence, was fed to a feed-forward network to generate a single label for the sequence. To localize the FOG episodes during evaluation, predictions for each sample were made by sliding the two-second partition in steps of one. This setting enables an objective comparison with the proposed action segmentation approaches as predictions are made at a temporal frequency of 100 Hz for both action detection schemes.

\subsection*{Evaluation}
\begin{table}[H]
\renewcommand{\arraystretch}{1.2}
\caption[Dataset characteristics.]{Overview of the number of motion capture trials (\#Trials), number of FOG trials (\#FOG trials), number of FOG episodes (\#FOG), percentage time spent frozen (\%TF), total duration of the FOG trials (in minutes), and average duration of the FOG trials ($\pm$ standard deviation (SD)) (in minutes). For dataset 1, the characteristics are given per subject. For dataset 2 (D2), which was only used to enrich the training dataset and not for model evaluation, a single summary is provided.\\}
\label{tab:data}
\centering
\footnotesize
\begin{tabular}{ccccccc}
\\
\hline
ID & \#Trials & \#FOG trials & \#FOG & \%TF & Total duration & Avg duration ($\pm$ SD) \\ \hline
S1 & 27       & 3            & 9     & 33.9 & 1.05 &  0.35 ($\pm$ 0.28)  \\
S2 & 22       & 11           & 13    & 12.3 & 4.01 &  0.36 ($\pm$ 0.12)  \\
S3 & 27       & 4            & 5     & 6.89 & 0.49 &  0.12 ($\pm$ 0.02)  \\
S4 & 21       & 9            & 18    & 36.7 & 2.45 &  0.27 ($\pm$ 0.19)  \\
S5 & 24       & 1            & 3     & 36.1 & 0.29 &  0.29  \\
S6 & 7        & 5            & 7     & 14.4 & 0.85 &  0.17 ($\pm$ 0.08)  \\
S7 & 31       & 1            & 1     & 20.1 & 0.08 &  0.08   \\
D2 & 68       & 68           & 134   & 28.4 & 24.4 &  0.36 ($\pm$ 0.18) \\ \hline
\end{tabular}
\end{table}

\begin{figure*}[!t]
\centering
\includegraphics[width=4.5in]{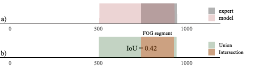}
\caption[Segment-wise loss: IoU and segment classification.]{Toy example to visualize the IoU computation and segment classification. The predicted FOG segmentation is visualized in pink, the experts' FOG segmentation in grey, and the colour gradient visualizes the overlap between the predicted and experts' segmentation. The intersection is visualized in orange and the union in green. If a FOG segment's IoU (intersection divided by union) crosses a predetermined threshold it is classified as a TP, if not, as a FP. For example, the FOG segment with an IoU of 0.42 would be classified as a FP. Given that the number of correctly detected segments (n=0) is less than the number of segments that the experts demarcated (n=1), there would be 1 FN.}
\label{fig:IoUjner}
\end{figure*}

For dataset 1, FOG was provoked for ten of the fourteen freezers during the test period, with seven subjects freezing within the visibility of the MoCap system. For dataset 2, eight of the seventeen freezers froze within the visibility of the MoCap system. The training dataset consists of the FOG and non-FOG trials of the seven subjects who froze in front of the MoCap system of dataset 1, enriched with the FOG trials of the eight subjects who froze in front of the MoCap system of dataset 2. Only the FOG trials of dataset 2 were considered to balance out the number of FOG and FG trials. Only the subjects of dataset 1 were considered for evaluation, as motivated in the procedure. Detailed dataset characteristics are provided in Table \ref{tab:data}. \\
The evaluation dataset was partitioned according to a leave-one-subject-out cross-validation approach. This cross-validation approach repeatedly splits the data according to the number of subjects in the dataset. One subject is selected for evaluation, while the other subjects are used to train the model. This procedure is repeated until all subjects have been used for evaluation. This approach mirrors the clinically relevant scenario of FOG assessment in newly recruited subjects \cite{Saeb2017-my}, where the model is tasked to assess FOG in unseen subjects. \\ 
From a machine learning perspective, action segmentation papers tend to use sample-wise metrics, such as accuracy, precision, and recall. However, sample-wise metrics do not heavily penalize over-segmentation errors. As a result, methods with significant qualitative differences, as was observed between the single-stage ST-GCN and MS-GCN, can still achieve similar performance on the sample-wise metrics. In 2016 Lea et al. \cite{Lea2016-yw} proposed a segment-wise F1-score to address those drawbacks. To compute the segment-wise F1-score, action segments are first classified as true positive (TP), false positive (FP), or false negative (FN) by comparing the intersection over union (IoU) to a pre-determined threshold, as visualized in Fig. \ref{fig:IoUjner}. The segment-wise F1-score has several advantages for FOG segmentation. (1) It penalizes over and under-segmentation errors, which would result in an inaccurate \#FOG severity outcome. (2) It allows for minor temporal shifts, which may have been caused by annotator variability and do not impact the FOG severity outcomes. (3) It is not impacted by the variability in FOG duration, since it is dependent on the number of FOG episodes and not on their duration. \\
This paper also reports a sample-wise metric. More specifically, the sample-wise Matthews correlation coefficient (MCC), defined as \cite{Matthews1975-hw}:
\begin{equation}
    MCC = \frac{TP*TN - FP*FN}{\sqrt{(TP+FP)(TP+FN)(TN+FP)(TN+FN)}}.
\end{equation}
A perfect MCC score is equal to one hundred, whereas minus one hundred is the worst value. An MCC score of zero is reached when the model always picks the majority class. The MCC can thus be considered a balanced measure, i.e. correct FOG and FG classification are of equal importance. 
The discrepancy between sample-wise MCC and the segment-wise F1 score allows assessment of potential over and under-segmentation errors. Conclusions were based on the segment-wise F1-score at high IoU overlap. \\
For the model validation, the entirety of dataset 1 was used, i.e. MoCap trials without FOG and MoCap trials with FOG, of the seven subjects who froze during the protocol. The machine learning metrics were used to evaluate MS-GCN with respect to the four strong baselines. While a high number of trials without FOG can inflate the metrics, correct classification of FOG and non FOG segments are, however, of equal importance for assessing FOG severity and thus also for assessing the performance of a machine learning model. To further assess potential false-positive scoring, an additional analysis was performed on trials without FOG of the healthy controls, non-freezers, and freezers that did not freeze during the protocol. \\
From a clinical perspective, FOG severity is typically assessed in terms of percentage time-frozen (\%TF) and number of detected FOG episodes (\#FOG) \cite{Morris2012-hl}. The \%TF quantifies the duration of FOG relative to the trial duration, and is defined as:
\begin{equation}
    \%TF = (\frac{1}{T} \sum_{t} y_{FOG}) * 100,
\end{equation}
where T are the number of samples in a MoCap trial and $y_{FOG}$ are the FOG samples predicted by the model or the samples annotated by the clinical experts. 
To evaluate the goodness of fit, the linear relationship between observations by the clinical experts and the model predictions was assessed. The strength of the linear relationship was classified according to \cite{Chan2003-pe}: $\geq 0.8$ : strong, $0.6 - 0.8$ : moderately strong, $0.3 - 0.5$ : fair, and $< 0.3$ : poor. The correlation describes the linear relationship between the experts' observations and the model predictions but ignores bias in predictions. Therefore, a linear regression analysis was performed to evaluate whether the linear association between the expert annotations and model predictions was statistically significant. The significance level for all tests was set at 0.05. For the FOG severity statistical analysis, only the trials with FOG were considered, as trials without FOG would inflate the reliability scores.

\section*{Results}
\subsection*{Model comparison}
\begin{table}[H]
\caption[Model comparison results.]{Overview of the FOG segmentation performance in terms of the segment-wise F1@50 and sample-wise MCC for MS-GCN and the four strong baselines. The $\dagger$ denotes the sliding window FOG detection scheme. All results were derived from the test set, i.e., subjects that the model had never seen.}
\label{tab:results_models}
\begin{adjustwidth}{-0.5in}{-0.5in}
\centering
\begin{tabular}{llllll}
\footnotesize
\\\hline
\multicolumn{1}{c}{Model} & \multicolumn{1}{c}{F1@10} & \multicolumn{1}{c}{F1@25} & \multicolumn{1}{c}{F1@50} & \multicolumn{1}{c}{F1@75} & \multicolumn{1}{c}{MCC} \\ \hline
Bi-LSTM$\dagger$ & 25.9 $\pm$ 8.40          & 21.8 $\pm$ 9.03          & 15.0 $\pm$ 5.60          & 11.9 $\pm$ 6.26          & 62.4 $\pm$ 23.2          \\
Bi-LSTM & 63.7 $\pm$ 21.7          & 63.2 $\pm$ 22.0          & 50.8 $\pm$ 25.4          & 40.9 $\pm$ 28.4          & 78.8 $\pm$ 21.1          \\
TCN     & 45.4 $\pm$ 16.8          & 42.7 $\pm$ 18.6          & 35.8 $\pm$ 14.8          & 27.0 $\pm$ 16.6          & 81.1 $\pm$ 12.9          \\
ST-GCN  & 53.2 $\pm$ 21.2          & 51.5 $\pm$ 21.7          & 46.7 $\pm$ 22.5          & 37.6 $\pm$ 26.6          & \textbf{83.0 $\pm$ 11.5} \\
MS-TCN  & 68.2 $\pm$ 29.4          & 66.8 $\pm$ 29.3          & 60.2 $\pm$ 30.5          & 54.9 $\pm$ 33.1          & 77.3 $\pm$ 22.2          \\
MS-GCN  & \textbf{77.8 $\pm$ 15.3} & \textbf{77.8 $\pm$ 15.3} & \textbf{74.2 $\pm$ 21.0} & \textbf{57.0 $\pm$ 30.1} & 82.7 $\pm$ 15.5       \\  \hline
\end{tabular}
\end{adjustwidth}
\end{table}

All models were trained using a leave-one-subject-out cross-validation approach. The metrics were summarized in terms of the mean $\pm$ standard deviation (SD) of the seven subjects that froze during the protocol, where the SD aims to capture the variability across different subjects. According to the results shown in table \ref{tab:results_models}, the ST-GCN-based models outperform the TCN and LSTM-based models on the MCC metric. This result confirms the notion that explicitly modelling the spatial hierarchy within the skeleton-based data results in a better representation \cite{Yan2018-jp}. Moreover, the multi-stage refinements improve the F1 score at all evaluated overlapping thresholds, the metric that penalizes over-segmentation errors, while the sample-wise MCC remains mostly consistent across stages. This result confirms the notion that multi-stage refinements can reduce the number of over-segmentation errors and improve neural network models for fine-grained activity segmentation \cite{Farha2019-yw}. Additionally, the results suggest that the sliding window scheme is ill-suited for fine-grained FOG annotation at high temporal frequencies.

\subsection*{MS-GCN detailed results}\label{qual1}
\begin{table}[H]
\renewcommand{\arraystretch}{1.2}
\caption[Detailed MS-GCN results.]{Detailed overview of the FOG assessment performance of the proposed MS-GCN model for each subject. The fourth column depicts the number of true positive FOG detections (TP) with respect to the number of FOG episodes. The fifth column depicts the number of false-positive (FP) FOG detections with respect to the number of trials that did not contain FOG. The sixth and seventh columns depict the \#FOG and \%TF computed from the model annotated segmentation with respect to those computed from the expert annotated segmentations. All results were derived from the test set, i.e., subjects that the model had never seen.}
\label{tab:results_MS-GCN}
\centering
\begin{tabular}{llllllllll}
\\\hline
\multicolumn{1}{c}{ID} & \multicolumn{1}{c}{F1@50} & \multicolumn{1}{c}{MCC} & \multicolumn{1}{c}{\#TP}  &  \multicolumn{1}{c}{\#FP}    & \multicolumn{1}{c}{\#FOG}   & \multicolumn{1}{c}{\%TF}        \\ \hline
S1 & 87.5  & 95.7 & 9 / 9 & 0 / 24  & 10 / 9  & 37.1 / 33.9 \\
S2 & 31.6  & 60.3 & 13 / 13 & 3 / 11  & 24 / 13 & 22.6 / 12.3 \\
S3 & 60.0  & 59.8 & 3 / 5 & 2 / 23  & 3 / 5   & 5.34 / 6.89 \\
S4 & 71.1  & 87.2 & 18 / 18 & 0 / 12  & 18 / 18 & 40.6 / 36.7 \\
S5 & 85.7  & 96.9 & 3 / 3 & 1 / 23  & 3 / 3   & 35.2 / 36.1 \\
S6 & 83.3  & 80.7 & 5 / 7 & 0 / 2   & 5 / 7   & 12.7 / 14.4 \\
S7 & 100   & 98.5 & 1 / 1& 0 / 30  & 1 / 1   & 19.5 / 20.1 \\ \hline
   & 74.2  & 82.7 & 52 / 56 & 6 / 125 & 64 / 56 & 24.7 / 22.9 \\ \hline
\end{tabular}
\end{table}

\begin{table}[H]
\renewcommand{\arraystretch}{1.2}
\caption[MS-GCN robustness.]{Overview of MS-GCN's robustness to false-positive FOG detections on the MoCap trials of the 14 healthy controls, 14 non-freezers, and 7 freezers that did not freeze in front of the cameras during the protocol. The letter k denotes the number of MoCap trials for each group.}
\label{tab:results_rob}
\centering
\begin{tabular}{lc}
\\\hline
Subjects & FP    \\\hline
Controls (k=404) & 0  \\
Non-freezers (k=423) & 0  \\
Freezers- (k=195) & 0  \\ \hline
\end{tabular}
\end{table}

\begin{figure*}[!t]
\centering
\includegraphics[width=4in]{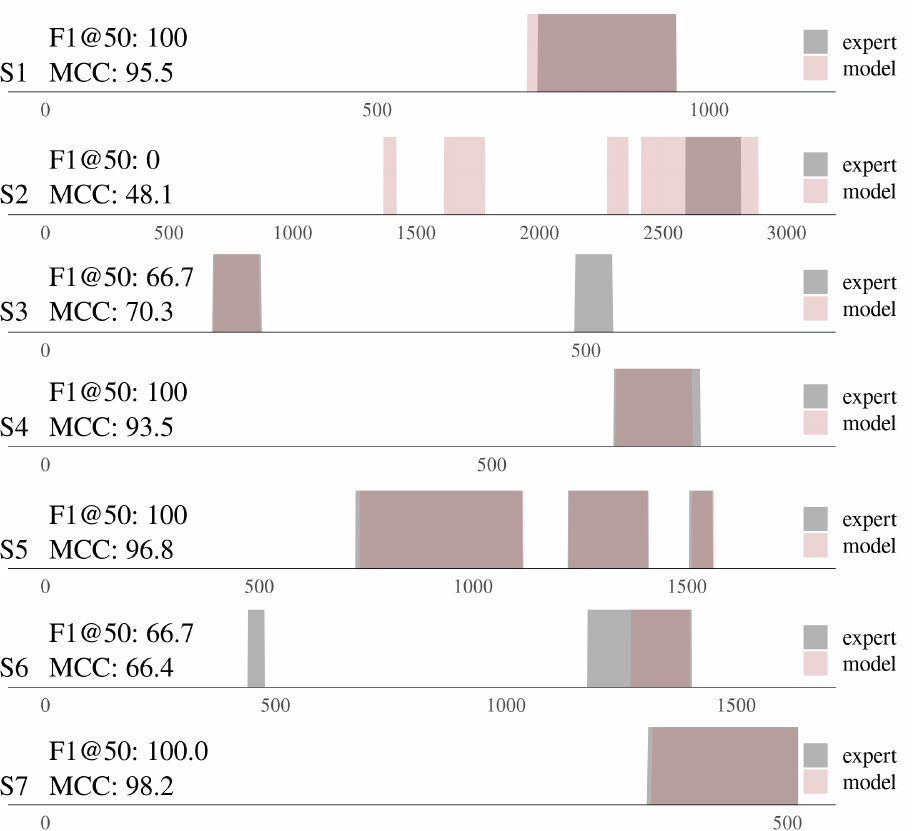}
\caption[Qualitative FOG segmentation results.]{Overview of seven standardized motion capture trials, visualizing the difference between the manual FOG segmentation by the clinician and the automated FOG segmentation by the MS-GCN. The x-axis denotes the number of samples (at a sampling frequency of 100 Hz). The colour gradient visualizes the overlap or discrepancy between the model and experts' annotations. The model annotations were derived from the test set, i.e., subjects that the model had never seen.}
\label{fig_qual_gcn}
\end{figure*}

This section provides an in-depth analysis of the performance of the MS-GCN model. According to the results shown in table \ref{tab:results_MS-GCN}, the model correctly detects 52 of 56 FOG episodes. A detection was considered as a TP if at least one sample overlapped with the ground-truth episode. Thus, without imposing a constraint on how much the predicted segment should overlap with the ground-truth segment, as is the case when computing the segment-wise F1 score. The model proved robust, with only six episodes incorrectly detected in a trial that the experts did not label as FOG. In terms of the clinical metrics, the model provides an accurate assessment of \#FOG and \%TF for five of the seven subjects. For S2 the model overestimates FOG severity, while for S3 the model underestimates FOG severity. \\ 
One FOG segmentation trial for each of the seven subjects is visualized in Fig. \ref{fig_qual_gcn}. The sample-wise MCC and segment-wise F1@50 for each trial are included for comparison. A near-perfect FOG segmentation can be observed for the trials of S1, S4, S5, and S7. For the two chosen trials of S3 and S6, the model did not detect two of the sub-0.5-second FOG episodes. For S2, it is evident that the model overestimates the number of FOG episodes. \\
A quantitative assessment of the MS-GCN predictions for the fourteen healthy control subjects (controls), fourteen non-freezers (non-freezers), and the seven freezers that did not freeze during the protocol (freezers-) further demonstrates the robustness of the MS-GCN. The results are summarized in table \ref{tab:results_rob}. According to table \ref{tab:results_rob}, no false-positive FOG segments were predicted. 

\subsection*{Automated FOG assessment: statistical analysis}
\begin{figure*}[!t]
\centering
\includegraphics[width=4in]{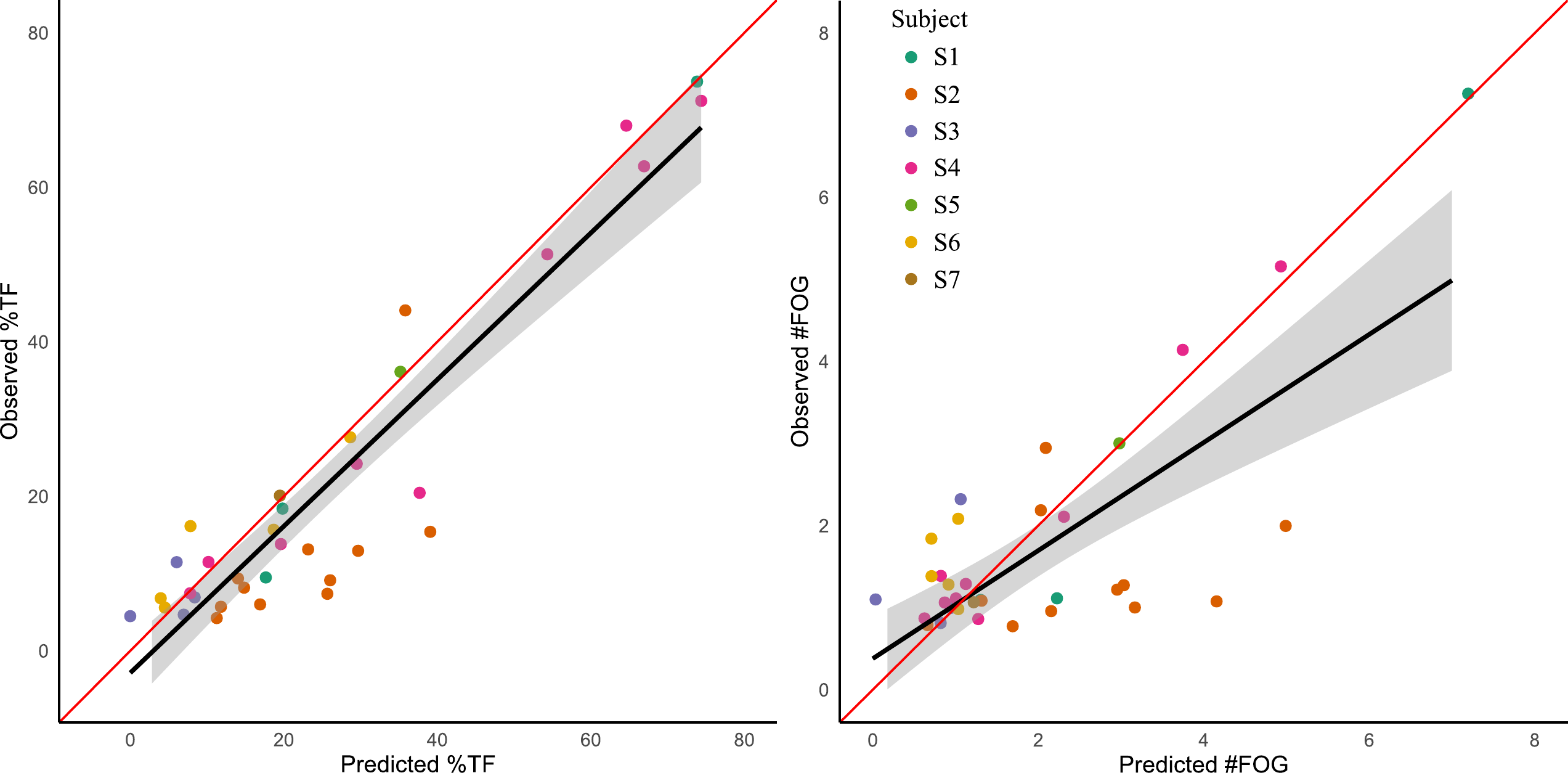}
\caption[FOG assessment: regression analysis.]{Assessing the performance of the MS-GCN (6 stages) for automated FOG assessment. More specifically, the performance to measure the percentage time-frozen (\%TF) (left) and the number of FOG episodes (\#FOG) (right) during a standardized protocol. The ideal regression line with a slope of one and an intercept of zero is visualized in red. All results were derived from the test set, i.e., subjects that the model had never seen. Observe the overestimation of \%TF and \#FOG for S2.}
\label{fig_rjner}
\end{figure*}

The clinical experts observed at least one FOG episode in 35 MoCap trials of dataset 1. The number of detected FOG episodes (\#FOG) per trial varied from 1 to 7 amounting to 56 FOG episodes, while the percentage time-frozen (\%TF) varied from 4.2 to 75. For the \%TF, the model predictions had a very strong linear relationship with the experts observations, with a correlation value [95 \% confidence interval (CI)] of r=0.93 [0.87, 0.97]. For the \#FOG, the model predictions had a moderately strong linear relationship with the experts' observations, with a correlation value [95 \% CI] of r=0.75 [0.55, 0.87]. A linear regression analysis was performed to evaluate whether the linear association between the experts' annotations and model predictions was statistically significant. For the \%TF, the intercept [95 \% CI] was -1.79 [-6.8, 3.3] and the slope [95 \% CI] was 0.96 [0.83, 1.1]. For the \#FOG, the intercept [95 \% CI] was 0.36 [-0.22, 0.94] and the slope [95 \% CI] was 0.73 [0.52, 0.92]. Given that the 95 \% CIs of the slopes exclude zero, the linear association between the model predictions and expert observations was statistically significant (at the 0.05 level) for both FOG severity outcomes. The linear relationship is visualized in figure \ref{fig_rjner}. 

\section*{Discussion}
Existing approaches treat automatic FOG assessment as an action recognition task and employ a sliding-window scheme to localize the FOG segments within a MoCap sequence. Such approaches require manually defined heuristics that may not generalize across study protocols. For instance, the most common FOG recognition scheme uses two-second partitions with majority voting to force all labels within a partition to a single label \cite{Sigcha2020-ft, ODay2021-ls}. Yet, such settings would induce a bias on the ground-truth annotations as sub-second episodes would never be the majority label. For the present dataset, this bias would neglect all the FOG episodes of S3. While shorter partitions could overcome this issue, they would restrict the amount of temporal context exposed to the model. \\
To address these issues, this paper reformulated FOG assessment as an action segmentation task. Action segmentation frameworks overcome the need for fixed partitioning by generating a prediction for each sample. Therefore, these frameworks rely only on the observations and their assumed model and not on manual heuristics that are unlikely to generalize across study protocols. As predictions vary at a high temporal frequency, action segmentation is inherently more challenging than recognition. To address this task, a novel neural network architecture, entitled MS-GCN, was proposed. MS-GCN extends MS-TCN \cite{Farha2019-yw}, the state-of-the-art model in action segmentation, to graph-based input data that is inherent to MoCap. \\
MS-GCN was quantitatively compared with four strong deep learning baselines. The comparison confirmed the notions that: (1) the multi-stage refinements reduce over-segmentation errors, and (2) the graph convolutions give a better representation of skeleton-based data than regular temporal convolutions. As a result, MS-GCN showed state-of-the-art FOG segmentation performance. 
Two common outcome measures to assess FOG, the \%TF and \#FOG  \cite{Morris2012-hl}, were computed and statistically assessed. MS-GCN showed a very strong (r=0.93) and moderately strong (r=0.75) linear relationship with the experts' observations for \%TF and \#FOG, respectively. For context, the intraclass correlation coefficient between independent assessors was reported to be 0.87 \cite{Walton2018-de} and 0.73 \cite{Morris2012-hl} for \%TF and 0.63 \cite{Morris2012-hl} for \#FOG. \\
A benefit of MS-GCN is that it is not strictly limited to marker-based MoCap data. The MS-GCN architecture naturally extends to other graph-based input data, such as single- or multi-camera markerless pose estimation \cite{Cao2018-yi, Mathis2018-pj}, and FOG assessment protocols that employ multiple on-body sensors \cite{Moore2013-ns, Popovic2010-jq}. Both technologies are receiving increased attention due to the potential to assess FOG not only in the lab but also in an at-home environment and thereby better capture daily-life FOG severity. Furthermore, up until now, deep learning-based gait assessment \cite{Kidzinski2019-ou, Kidzinski2020-ad, Lempereur2020-at, Filtjens2020-hl} did not yet exploit the inherent graph-structured data. The established improvement in FOG assessment by this research might, therefore, signify further improvements in deep learning-based gait assessment in general. \\
Several limitations are present. The first and most prominent limitation is the lack of variety in the standardized FOG-provoking protocol. FOG is characterized by several apparent subtypes, such as turning and destination hesitation, and gait initiation \cite{Schaafsma2003-pz}. While turning was found to be the most prominent \cite{Schaafsma2003-pz, Giladi2002-bf}, it should still be established whether MS-GCN can generalize to other FOG subtypes under different FOG provoking protocols. For now, practitioners are advised to closely follow the experimental protocol used in this study when employing MS-GCN. The second limitation is the small sample size. While MS-GCN was evaluated based on the clinically relevant use-case scenario of FOG assessment in newly recruited subjects, the sample size of the dataset is relatively small compared to the deep learning literature. The third limitation is based on the observation that FOG assessment in the clinic and lab is prone to two shortcomings. (1) FOG can be challenging to elicit in the lab due to elevated levels of attention \cite{Snijders2008-vt, Okuma2014-tl}, despite providing adequate FOG provoking circumstances \cite{Spildooren2010-pj, Nieuwboer2001-cr}. (2) Research has demonstrated that FOG severity in the lab is not necessarily representative of FOG severity in daily life \cite{Rahman2008-rg, Snijders2008-vt}. Future work should therefore establish whether the proposed method can generalize to tackle automated FOG assessment with on-body sensors or markerless MoCap captured in less constrained environments. Fourth, due to the opaqueness inherent to deep learning, clinicians have historically distrusted DNNs \cite{Barredo_Arrieta2020-gd}. However, prior case studies \cite{Horst2019-zo, Filtjens2021-tp}, have demonstrated that interpretability techniques are able to visualize what features the model has learned \cite{Bach2015-gm, Sundararajan2017-uh, Shrikumar2017-au}, which can aid the clinician in determining whether the assessment was based on credible features. 

\section*{Conclusion}
FOG is a debilitating motor impairment of PD. Unfortunately, our understanding of this phenomenon is hampered by the difficulty of objectively assessing FOG. To tackle this problem, this paper proposed a novel deep neural network architecture. The proposed architecture, termed MS-GCN, was quantitatively validated versus the expert clinical opinion of two independent raters. In conclusion, it can be established that MS-GCN demonstrates state-of-the-art FOG assessment performance. Furthermore, future work is now possible that aims to assess the generalization of MS-GCN to other graph-based input data, such as markerless MoCap or multiple on-body sensor configurations, and to other FOG subtypes captured under less constrained protocols. Such work is important to increase our understanding of this debilitating phenomenon during everyday life.



\begin{backmatter}
\section*{List of abbreviations}
FOG: Freezing of Gait; PD: Parkinson's Disease; PwPD: People with Parkinson's Disease; \%TF: Percentage Time Spent Frozen; \#FOG: Number of FOG Episodes; MoCap: Motion Capture; TCN: Temporal Convolutional Neural Network; MS-TCN: Multi-Stage Temporal Convolutional Neural Network; GCN: Graph Convolutional Neural Networks; ST-GCN: Spatial-Temporal Graph Convolutional Neural Network; MS-GCN: Multi-Stage Spatial-Temporal Graph Convolutional Neural Network; NFOG-Q: New Freezing of Gait Questionnaire;  H\&Y: Hoehn and Yahr; MMSE: Mini-Mental State Examination; UPDRS: Unified Parkinson's Disease Rating Scale; SD: Standard Deviation; D2: Dataset 2; FG: Functional Gait; TP: True Positive; TN: True Negative; FP: False positive; FN: False Negative; MCC: Matthews Correlation Coefficient; CI: Confidence Interval; BTK: Biomechanical Toolkit.

\section*{Declarations}
\subsection*{Ethics approval and consent to participate}
The study was approved by the local ethics committee of the University Hospital Leuven and all subjects gave written informed consent.

\subsection*{Consent for publication}
Not applicable

\subsection*{Availability of data and materials}
The input set was imported and labeled using Python version 2.7.12 with Biomechanical Toolkit (BTK) version 0.3 \cite{Barre2014-os}. The MS-GCN architecture was implemented in Pytorch version 1.2 \cite{Paszke2019-bz} by adopting the public code repositories of MS-TCN \cite{Farha2019-yw} and ST-GCN \cite{Yan2018-jp}. All models were trained on an NVIDIA Tesla K80 GPU using Python version 3.6.8. The datasets analyzed during the current study are not publicly available due to restrictions on sharing subject health information. 

\subsection*{Competing interests}
The authors declare that there is no conflict of interest regarding the publication of this article.
  
\subsection*{Funding}
This research did not receive any specific grant from funding agencies in the public, commercial, or not-for-profit sectors.

\subsection*{Author's contributions}
Study design by BF, PG, AN, PS, and BV. Data analysis by BF. Design and implementation of the neural network architecture by BF. Statistics by BF and BV. Subject recruitment, data collection, and data preparation by AN. The first draft of the manuscript was written by BF and all authors commented on subsequent revisions. The final manuscript was read and approved by all authors.

\subsection*{Acknowledgements}
We thank the employees of the gait laboratory for technical support during data collection.


\bibliographystyle{bmc-mathphys} 
\bibliography{bmc_article}      

\end{backmatter}
\end{document}